%% file: paper_iswc2012.tex
\documentclass{llncs}
\usepackage{makeidx}  % allows for indexgeneration
\usepackage{times}
\usepackage{graphicx}
\usepackage{latexsym,footnote}
\usepackage{amsmath}
\usepackage[ruled,vlined,linesnumbered,norelsize]{algorithm2e}
\usepackage{mathnotation}

\usepackage[T1]{fontenc} 
\usepackage{color}
\usepackage[draft]{fixme}

\usepackage{subfigure}

\fxsetup{
nomargin,inline,index,
theme=color
}

\newcommand{\tax}{\mathit{ax}}

\newcommand{\mo}{\mathcal{O}}
\newcommand{\mt}{\mathcal{T}}
\newcommand{\ma}{\mathcal{A}}

\newcommand{\mb}{\mathcal{B}}
\newcommand{\md}{\mathcal{D}}

\newcommand{\Tp}{\mathit{P}}
\newcommand{\Tn}{\mathit{N}}
\newcommand{\tp}{\mathit{p}}
\newcommand{\tn}{\mathit{n}}
\newcommand{\dt}{\mathcal{D}_{t}}

\newcommand{\ot}{\mathcal{O}_{t}}

\newcommand{\mX}{{\bf{X}}}

\newcommand{\mD}{{\bf{D}}}
\newcommand{\dx}[1]{{\bf D}_{#1}^P}
\newcommand{\dnx}[1]{{\bf D}_{#1}^{N}}
\newcommand{\dz}[1]{{\bf D}_{#1}^\emptyset}
 
\newcommand{\qc}{\mathit{caut}}
\newcommand{\uc}{c}

\newcommand{\Align}{{\mathit{M}}}

\newcommand{\HR}{{\mathit{HR}}}

\newcommand{\ax}{{\mathit{ax}}}

\newcommand{\RQ}{{\mathit{R}}}

\newcommand{\Xsc}{\mathit{X_{sc}}}
\newcommand{\Xalt}{\mathit{X_{alt}}}

\begin{document}
%titlenote
\title{RIO: Minimizing User Interaction \\in Ontology Debugging}
%\title{RIO: A Risk Optimization Approach for Interactive Ontology Debugging}
\author{Patrick Rodler \and Kostyantyn Shchekotykhin \and Philipp Fleiss \and Gerhard Friedrich }
\authorrunning{Patrick Rodler et al.} % abbreviated author list (for running head)
\tocauthor{Patrick Rodler, Kostyantyn Shchekotykhin, Philipp Fleiss, Gerhard Friedrich}
\institute{Alpen-Adria Universit\"at, Klagenfurt, 9020 Austria \\ \email{firstname.lastname@aau.at}}

\maketitle
\bibliographystyle{splncs03}

\vspace{-10pt}
\begin{abstract}
Efficient ontology debugging is a cornerstone for many activities in the context of the Semantic Web, especially when automatic tools produce (parts of) ontologies such as in the field of ontology matching. The best currently known interactive debugging systems rely upon some meta information in terms of fault probabilities, which can speed up the debugging procedure in the good case, but can also have negative impact on the performance in the bad case. The problem is that assessment of the meta information is only possible a-posteriori. Consequently, as long as the actual fault is unknown, there is always some risk of suboptimal interactive diagnoses discrimination. As an alternative, one might prefer to rely on a tool which pursues a no-risk strategy. In this case, however, possibly well-chosen meta information cannot be exploited, resulting again in inefficient debugging actions. In this work we present a reinforcement learning strategy that continuously adapts its behavior depending on the performance achieved and minimizes the risk of using low-quality meta information. Therefore, this method is suitable for application scenarios where reliable a-priori fault estimates are difficult to obtain. 
Using problematic ontologies in the field of ontology matching, we show that the proposed risk-aware query strategy outperforms both active learning approaches and no-risk strategies on average in terms of required amount of user interaction.
\end{abstract}

%147 diagnoses
%- extreme good: 9 queries (all diagnoses are similar - algorithm cannot distinguish between them - therefore bad performance although 'extreme good')
%- split: 6 queries

%%%%%%%%%%%%%%%%%%%%%%%%%%%%%%%%%%%%%%%%%%%%%%%%%%%
%Nummerierung bei equations entfernen, die nie referenziert werden
%%%%%%%%%%%%%%%%%%%%%%%%%%%%%%%%%%%%%%%%%%%%%%%%%%%
%\vspace{-20pt}
\input{intro}
\input{basic}
\input{strategy}
\input{eval}

\vspace{-20pt}
\section{Conclusion}
\label{sec:conclusion}
\vspace{-5pt}
We have shown problems of state-of-the-art interactive ontology debugging strategies w.r.t.\ the usage of unreliable meta information. To tackle this issue, we proposed a learning strategy which combines the benefits of existing approaches, i.e. high potential and low risk. Depending on the performance of the diagnosis discrimination actions, the trust in the a-priori information is adapted. Tested under various conditions, our algorithm 
%required minimal amount of user interaction on average in all experiments
revealed an average performance superior to two common approaches in the field w.r.t. required user interaction. 
%independently of the given meta information.
In our evaluation we showed the utility of our approach in the important area of ontology matching, its scalability and adequate reaction time allowing for continuous interactivity. 
%Our method enables to search for faults in the most general debugging setting comprising not only the alignment, but also the matched ontologies, where the a-priori fault information might be unreliable. 

%\section{Acknowledgements}
%Special thanks to Christian Meilicke for the supply of test cases used in the evaluation.

%In this paper we presented a risk optimization approach to sequential debugging of ontologies. The proposed method exploits meta information in terms of fault probabilities and user-defined willingness to take risk. A substantial cost reduction compared to existing methods was shown for any quality of the given meta information. The proposed method is of particular significance for domains where only vague meta information is available, such as ontology development.   
%
%advantages of existing methods
%entropy-based query selection and split-in-half. 
%is designed to stabilize the performance of entropy-based query selection in case the target diagnosis is not favored by the prior fault probabilities. 
%For these cases a significant cost reduction was demonstrated.  

%We showed that performance of the entropy-based approach given only weakly justified priors can be significantly improved by limiting the risk that the query selection is allowed to take. ...

%\includegraphics[width=\textwidth]{evaluation_iswc2011.pdf}\label{fig:eval}

\bibliography{V-Know}
\end{document}

%% file: intro.tex
\section{Introduction} 
%Support of ontology development and maintenance is an important requirement for the extensive use of Semantic Web technologies. 

%%%%%%%%%%%%%%%%%%%%
%Approach allows to minimize user interaction in interactive ontology debugging on average.
%%%%%%%%%%%%%%%%%%%%%%

The foundation for widespread adoption of Semantic Web technologies is a broad community of ontology developers which is not restricted to experienced knowledge engineers. Instead, domain experts from diverse fields should be able to create ontologies incorporating their knowledge as autonomously as possible. The resulting ontologies are required to fulfill some minimal quality criteria, usually consistency, coherency and no undesired entailments, in order to grant successful deployment. However, 
%this brings about the following problems: (1) 
the correct formulation of logical descriptions in ontologies is an error-prone task which accounts for a need for assistance in ontology development in terms of ontology debugging tools. Usually, such tools~\cite{schlobach2007,Kalyanpur.Just.ISWC07,friedrich2005gdm,Horridge2008} use model-based diagnosis~\cite{Reiter87} to identify sets of faulty axioms, called diagnoses, that need to be modified or deleted in order to meet the imposed quality requirements. The major challenge inherent in the debugging task is often a substantial number of alternative diagnoses. This problem has been addressed in \cite{jws12} by proposing a debugging method based on active learning which exploits additional information in terms of queries to a user about the intended ontology. Thereby, the selection of queries is guided by the specification of some meta information, i.e. prior knowledge about fault probabilities of a user w.r.t. particular logical operators. When chosen appropriately, this meta information proved to be very useful in that the interaction with the domain expert can be drastically reduced. However, given that only poor prior knowledge is available, the amount of interaction increased compared to methods which manifest constant performance without taking into account any meta information.

A similar interactive technique can be found in~\cite{Nikitina2011}, where queries to a user are incorporated to revise an ontology. Ontology revision aims at partitioning a given ontology into a set of correct axioms and a set of incorrect ones. The system can deal with inconsistent/incoherent ontologies only after a union of all axioms causing these problems is identified and added to the initial set of incorrect axioms. Computation of these axioms, however, requires ontology debugging, which is not addressed in the paper.
% do not rely upon meta information at all. 

In a debugging scenario involving 
%only a single 
a faulty ontology developed by one expert, the meta information might be extracted from the logs of previous sessions, if available, or specified by the expert based on their experience w.r.t. own faults. However, in scenarios involving automatized systems producing (parts of) ontologies, e.g. ontology alignment and ontology learning, or numerous users collaborating in modeling an ontology, the choice of reasonable meta information is rather unclear.
%such meta information is usually not available. Then it is 
%in the situation of, e.g., ontology alignment or ontology learning, such meta information is usually not available.
%In a debugging scenario involving only a single faulty ontology developed by one expert, the meta information might be extracted from the logs of previous sessions or specified by the expert based on their experience w.r.t. own faults. However, in the situation of collaborative development and ontology alignment where multiple independent and autonomously built ontologies describing related domains are joined by means of automatized ontology matching tools, the choice of reasonable meta information is rather unclear. Usually, the single constituents of the aligned ontology are correct when considering each of them as a separate description of its domain. Only their integration by adding mapping correspondences between concepts and properties, respectively, may cause violation of quality requirements. Yet this does not necessarily mean that the fault consists solely of newly added mapping axioms. In fact, there is no information at all about the actual location of the fault, which may be (1) only in the mapping, (2) only in some ontologies, or (3) distributed over the mapping and some ontologies. Current alignment diagnosis tools, e.g.~\cite{Meilicke2008a}, do not tackle this general problem, but address only case~(1). 
If, on the one hand, an active learning method is used relying on a guess of the meta information, this might result in an overhead w.r.t. user interaction of more than 2000\%. If one wants to play it safe, on the other hand, by deciding not to exploit any meta information at all, this might also result in substantial extra time and effort for the user. So, thitherto one is spoilt for choice between strategies with high potential but also high risk, or methods with no risk but also no potential.

In this work we present an ontology debugging approach with high potential and low risk, which allows to minimize user interaction throughout a debugging session on average, without depending on high-quality meta information. 
%provides a solution for (1), (2) and (3) in the ontology alignment case and 
By virtue of its reinforcement learning capability, our approach is optimally suited for debugging ontologies where only vague or no meta information is available. On the one hand, our method takes advantage of the given meta information as long as good performance is achieved. On the other hand, it gradually gets more independent of meta information if suboptimal behavior is measured. Moreover, our strategy can take into account an expert's subjective quality estimation of the meta information. In this way an expert may decide to take influence on the algorithm's behavior by limiting the range of admissible values the learning parameter may take. Alternatively, the algorithm acts freely and finds a profitable strategy on its own.
%. To find the best strategy, our method 
This is accomplished by constantly improving the quality of meta information and adapting a risk parameter based on the new information obtained by queries answered by the user.
%new information gathered by means 
%of queries. 
%as well as it learns 
%to act profitably 
%in all kinds of situations 
%by exploiting its adaptiveness. 
This means that, in case of good meta information, the performance of our method will be close to the performance of the active learning method, whereas, in case of bad meta information, the achieved performance will approach the performance of the risk-free strategy. So, our approach can be seen as a risk optimization strategy (RIO) which combines the benefits of active learning and risk-free strategies. Experiments on two datasets of faulty ontologies
%mappings produced by automatic ontology alignment tools 
show the feasibility, efficiency and scalability of RIO. The evaluation of these experiments will manifest that, on average, RIO is the best choice of strategy for both good and bad meta information with savings in terms of user interaction of up to 80\%.

The problem specification, basic concepts and a motivating example are provided in Section~\ref{sec:basics}. Section~\ref{sec:theory} explains the suggested approach and gives implementation details. Evaluation results are described in Section~\ref{sec:eval}. Section \ref{sec:conclusion} concludes.

%% file: basic.tex
\section{Basic Concepts and Motivation} \label{sec:basics}
Ontology debugging deals with the following problem: Given is an ontology $\mo$ which does not meet postulated requirements $\RQ$, e.g. $\RQ=\{\text{coherency},\text{consistency}\}$. $\mo$ is a set of axioms formulated in some monotonic knowledge representation language, e.g. OWL.
%, which does not meet postulated requirements $\RQ$, e.g. $\RQ=\{\text{coherency},\text{consistency}\}$. 
The task is to find a subset of axioms in $\mo$, called diagnosis, that needs to be altered or eliminated from the ontology in order to meet the given requirements. 
To this end, our approach to ontology debugging presumes sound and complete procedures for deciding logical consistency and for calculating logical entailments, which are used as a black box. For OWL, e.g., both functionalities are provided by a standard DL-reasoner.

Generally, there are many diagnoses for one and the same faulty ontology $\mo$. The problem is then to figure out the single diagnosis, called target diagnosis $\dt$, 
that complies with the knowledge to be modeled by the intended ontology.
%facts intended to be modeled by the ontology. 
In interactive ontology debugging we assume a user, e.g. the author of the faulty ontology or a domain expert, interacting with an ontology debugging system by answering queries about entailments of the desired ontology, called the target ontology $\ot$. The target ontology can be understood as $\mo$ minus the axioms of $\dt$ plus additional axioms $EX_{\dt}$ which can be added in order to regain desired entailments which might have been eliminated together with axioms in $\dt$. Note that the user is not expected to know $\ot$ explicitly (in which case there would be no need to consult an ontology debugger), but implicitly in that they are able to answer queries about $\ot$. 
Roughly speaking, each query is a set of logical descriptions and the user is queried whether the conjunction of these descriptions is entailed by $\ot$. Every positively (negatively) answered query constitutes a positive (negative) test case fulfilled by $\ot$. The set of positive (entailed) and negative (non-entailed) test cases is denoted by $\Tp$ and $\Tn$, respectively. So, $\Tp$ and $\Tn$ are sets of sets of axioms, which can be, but do not need to be, initially empty. Test cases can be seen as constraints $\ot$ must satisfy and are therefore used to gradually reduce the search space for valid diagnoses. Simply put, the overall procedure consists of (1)~computing a predefined number of diagnoses, (2)~gathering additional information by querying the user, (3)~incorporating this information to cut irrelevant areas off the search space, and so forth, until the search space is reduced to a single (target) diagnosis $\dt$. 

The general debugging setting we consider also envisions the opportunity for the user to specify some background knowledge $\mb$, i.e.~a set of axioms which are known to be correct.
%, which is 
$\mb$ is then incorporated in the calculations throughout the ontology debugging procedure. 
For example, in case the user knows that a subset of axioms in $\mo$ is definitely sound, all axioms in this subset are added to $\mb$ before initiating the debugging session. Then, $\mb$ and $\mo~\setminus~\mb$ partition the original ontology into a set of correct and possibly incorrect axioms, respectively. In the debugging session, only $\mo := \mo \setminus \mb$ is used to search for diagnoses. This can reduce the search space for diagnoses substantially.

More formally, ontology debugging can be defined in terms of conditions a target ontology must fulfill, which leads to the definition of a diagnosis problem instance, for which we search for solutions, i.e. diagnoses:
\begin{definition}[Target Ontology, Diagnosis Problem Instance] Let $\mo = (\mt,\ma)$ denote an ontology consisting of a set of terminological axioms $\mt$ and a set of assertional axioms $\ma$, $\Tp$ a set of positive test cases, $\Tn$ a set of negative test cases, $\mb$ a set of background knowledge axioms, and $\RQ$ a set of requirements to an ontology\footnote{Throughout the paper we consider debugging of inconsistent and/or incoherent ontologies, i.e. whenever not stated explicitly we assume $\RQ=\{\text{consistency, coherency}\}$. 
%However, the same approach can also be used with other requirements such as satisfiability of concepts (coherency) that can be formulated instead or in addition to consistency.
}.
Then an ontology $\ot$ is called \emph{target ontology} iff all the following conditions are fulfilled:
\vspace{-3pt}
\begin{eqnarray*}
		 \forall \, r  \in \RQ&:& \;\ot \cup \mb \,\text{ fulfills }\, r  \\
		 \forall \,\tp \in \Tp&:& \;\ot \cup \mb \,\models\, \tp				\\
		 \forall \,\tn \in \Tn&:& \;\ot \cup \mb \,\not\models\, \tn 
\end{eqnarray*}
\vspace{-14pt}\\
The tuple $\langle\mo,\mb,\Tp,\Tn\rangle_\RQ$ is called a \emph{diagnosis problem instance} iff $\mb \cup (\bigcup_{\tp \in \Tp} \tp) \not\models \tn$ for all $\tn \in \Tn$ and $\mo$ is not a target ontology, i.e. $\mo$ violates at least one of the conditions above.
%following 
%conditions defining a target ontology. 
%statements (1),(2),(3) is true: (1)~$\exists \, r  \in \RQ: \, \mo \cup \mb \text{ violates } r$,\footnote{Throughout the paper we consider debugging of inconsistent ontologies, i.e. whenever not stated eplicitly we assume $\RQ=\{\text{consistency}\}$. However, the same approach can also be used with other requirements such as satisfiability of concepts (coherency) that can be formulated instead or in addition to consistency.} (2)~$\exists \,\tp \in \Tp: \, \mo \cup \mb \not\models \tp$, (3)~$\exists \,\tn \in \Tn: \, \mo \cup \mb \models \tn$. 
%\begin{eqnarray*}	
%\exists \, r  \in \RQ&:& \; \mo \cup \mb \text{ violates } r\footnote{Throughout the paper we consider debugging of inconsistent ontologies, i.e. whenever not stated eplicitly we assume $\RQ=\{\text{consistency}\}$. However, the same approach can also be used with other requirements such as satisfiability of concepts (coherency) that can be formulated instead or in addition to consistency.} \\
%\exists \,\tp \in \Tp&:& \; \mo \cup \mb \not\models \tp   \\
%\exists \,\tn \in \Tn&:& \; \mo \cup \mb \models \tn 
%\end{eqnarray*}
\end{definition}

\begin{definition}[Diagnosis]\label{def:diagnosis}
We call $\md \subseteq \mo$ a \emph{diagnosis} w.r.t. a diagnosis problem instance $\langle\mo,\mb,\Tp,\Tn\rangle_\RQ$ iff there exists a set of axioms $EX_\md$ such that 
$(\mo\setminus\md)\cup EX_\md$ 
%meets all the conditions defining 
is a target ontology.
%all the following conditions hold:
%	\begin{eqnarray*}
%		\label{req_1} \forall \, r  \in \RQ&:& \;(\mo\setminus\md)\cup \mb\cup EX_\md \,\text{ fulfills }\, r  \\
%		\label{req_2} \forall \,\tp \in \Tp&:& \;(\mo\setminus\md)\cup \mb\cup EX_\md \,\models\, \tp				\\
%		\label{req_3} \forall \,\tn \in \Tn&:& \;(\mo\setminus\md)\cup \mb\cup EX_\md \,\not\models\, \tn
%	\end{eqnarray*} 
A diagnosis $\md$ is \emph{minimal} iff there is no $\md' \subset \md$ such that \ $\md'$ is a diagnosis. A diagnosis $\md$ gives complete information about the correctness of each axiom $ax_k \in \mo$, i.e. all $ax_i \in \md$ are assumed to be faulty and all $ax_j \in \mo\setminus\md$ are assumed to be correct. The set of all minimal diagnoses is denoted by $\mD$.
\end{definition}
The identification of an extension $EX_\md$, accomplished e.g. by some learning approach, is a crucial part of the ontology repair process. However, the formulation of a complete extension is outside the scope of this work where we focus on computing diagnoses. Following the approach suggested in~\cite{jws12}, we approximate $EX_\md$ by the set $\bigcup_{\tp \in \Tp} \tp$.

An immediate consequence of Definition \ref{def:diagnosis} is: The more test cases are specified, the fewer minimal diagnoses $\mD$ exist for a diagnosis problem instance. So, the uncertainty about the target diagnosis $\dt \in \mD$ is gradually reduced by specifying test cases. 

\noindent\textbf{Example:} 
%Assume an expert wants to match two consistent ontologies $\mo_1$ and $\mo_2$, which describe PhD students as researchers and as university students correspondingly. 
Consider the OWL ontology $\mo$ encompassing the following terminology $\mt$:
\begin{center}
\vspace{-2pt}
\begin{tabular}{crl}
        & $\tax_1:$ &  $PhD \sqsubseteq Researcher$ \\ 
        & $\tax_2:$ &  $Researcher \sqsubseteq DeptEmployee$ \\ 
        & $\tax_3:$ &  $PhDStudent \sqsubseteq Student$  \\ 
        & $\tax_4:$ &  $Student \sqsubseteq \lnot DeptMember$ \\ 
        & $\tax_5:$ &  $PhDStudent \sqsubseteq PhD$ \\
        & $\tax_6:$ &  $DeptEmployee \sqsubseteq DeptMember$
\end{tabular}
\vspace{-2pt}
\end{center}
and an assertional axiom $\ma=\setof{PhDStudent(s)}$. Then $\mo$ is inconsistent since it describes a PhD student as both a department member and not. 

Let us assume that the assertion $PhDStudent(s)$ is considered as correct and is thus added to the background theory, i.e. $\mb = \ma$, and both sets $\Tp$ and $\Tn$ are empty. Then, the set of minimal diagnoses $\mD=\{\md_1 : [\tax_1], \md_2 : [\tax_2], \md_3 : [\tax_3], \md_4 : [\tax_4], \md_5 : [\tax_5],\md_6 : [\tax_6]\}$ for the given problem instance $\tuple{\mt,\ma,\emptyset,\emptyset}$. $\mD$ can be computed by a diagnosis algorithm such as the one presented in~\cite{friedrich2005gdm}. 

With six diagnoses for six ontology axioms, this example might already give an idea that in many cases the number of diagnoses $\mD$ can get very large. Without any prior knowledge, each of the diagnoses in $\mD$ is equally likely to be the target diagnosis $\dt$. So, it depends on the specified test cases, i.e.~answers to the queries asked to the user, which diagnosis will be the target diagnosis. The test cases, however, represent properties, i.e. entailments and non-entailments, of the target ontology $\ot:=(\mo \setminus \dt) \cup EX_{\dt}$ and thus allow to constrain the possibilities for $\dt$.
In order to define a query~\cite{jws12}, the fact is exploited that ontologies $\mo\setminus\md_i$ and $\mo\setminus\md_j$ resulting in application of different diagnoses $\md_i,\md_j \in \mD$\, ($\md_i \neq \md_j$) entail different sets of logical descriptions. When we speak of entailments, we address the output computed by the classification and realization services of a reasoner. Formally, a query is defined as follows:
\begin{definition}[Query]
A set of logical descriptions $X_j$ is called a \emph{query} iff there exists a set of diagnoses $\emptyset \subset \mD' \subset \mD$ such that $X_j$ is entailed by each ontology in $\setof{\mo^{*}_i\,|\,\md_i \in \mD'}$ where $\mo^{*}_i := (\mo \setminus \md_i) \cup \mb \cup \bigcup_{\tp\in\Tp} \tp$. Asking a query $X_j$ to a user means asking them $(\ot \models X_j ?)$. The set of all queries w.r.t. $\mD$ is denoted by $\mX_\mD$.\footnote{For the sake of simplicity, we will use $\mX$ instead of $\mX_\mD$ throughout this work because the $\mD$ associated with $\mX$ will be clear from the context.}
\end{definition}
Each query $X_j$ partitions the set of diagnoses $\mD$ into $\langle \dx{j}, \dnx{j}, \dz{j} \rangle$ such that:
\vspace{-7pt}
\begin{eqnarray*}\label{partition}
	\dx{j} &=& \setof{ \md_i \;|\; \mo^{*}_i \,\models\, X_j} \\
	%\cup \mb \cup \bigcup_{\tp\in\Tp} \tp
	\dnx{j} &=& \setof{ \md_i \;|\; \mo^{*}_i \cup X_j \text{ is inconsistent}}\\
	\dz{j} &=& \mD \setminus (\dx{j} \cup \dnx{j})
\end{eqnarray*}
\vspace{-17pt}\\
%Consequently, if the oracle answers \emph{yes} to the query $X_j$, then we can add the query to the positive test cases $\Tp \cup \setof{X_j}$ and \emph{reject} all diagnoses in $\dnx{j}$, otherwise given an answer \emph{no} the query is added to the negative test cases $\Tn \cup \setof{X_j}$ and all diagnoses in $\dx{j}$ are \emph{rejected}.
%Consequently, if the oracle answers \emph{yes} to the query $X_j$, then $X_j$ is added to the positive test cases, i.e. $\Tp \leftarrow \Tp \cup \setof{X_j}$, and all diagnoses in $\dnx{j}$ are \emph{rejected}. Given the answer \emph{no}, $\Tn \leftarrow \Tn \cup \setof{X_j}$ and all diagnoses in $\dx{j}$ are \emph{rejected}.
If the answering of queries by a user $u$ is modeled as a function $a_u: \mX \rightarrow \{\textit{yes},\textit{no}\}$, then the following holds: If $a_u(X_j) = \textit{yes}$, then $X_j$ is added to the positive test cases, i.e. $\Tp \leftarrow \Tp \cup \setof{X_j}$, and all diagnoses in $\dnx{j}$ are \emph{rejected}. Given that $a_u(X_j) = \textit{no}$, then $\Tn \leftarrow \Tn \cup \setof{X_j}$ and all diagnoses in $\dx{j}$ are \emph{rejected}.

This allows us to formulate the subproblem of ontology debugging addressed in this work:
%However, it is still unclear which of these diagnoses corresponds to the target diagnosis $\md_t$ that allows the formulation of the intended matching result $\mo_t$.
%Moreover, in many cases an ontology debugger can return a huge set of diagnoses $\mD$ making the discrimination between the diagnoses even more complicated, if not impossible. To tackle this problem, the authors in~\cite{jws12} exploit the fact that ontologies $\mo\setminus\md_i$ and $\mo\setminus\md_j$ resulting in application of different diagnoses $\md_i,\md_j \in \mD$\, ($\md_i \neq \md_j$) entail different sets of logical descriptions. These entailments, computed by the classification and realization services of a reasoner, can be used to query an oracle (e.g. a domain expert or an information system). The answers can be used to discriminate between the diagnoses.
\vspace{-2pt}
\begin{definition}[Diagnosis Discrimination]
\textbf{Given} the set of diagnoses $\mD = \{\md_1,\dots,\md_n\}$ w.r.t. $\langle\mo,\mb,\Tp,\Tn\rangle_\RQ$ and a user $u$, \textbf{find} a sequence $(X_1,\dots,X_q)$ of queries $X_i \in \mX$ with minimal $q$, such that $\mD =\{\dt\}$ after assigning 
$X_{i(i=1\dots,q)}$ each to either $\Tp$ iff $a_u(X_i) = \textit{yes}$ or $\Tn$ iff $a_u(X_i) = \textit{no}$.\footnote{Since the user $u$ is assumed fixed throughout a debugging session and for brevity, we will use $a_i$ equivalent to $a_u(X_i)$ in the rest of this work.}
\end{definition}
\vspace{-2pt}
%Then the approximated target diagnosis $\ot :=$ is then defined
%$X_1,\dots,X_q$ each to either $\Tp$ (in case of positive answer) or $\Tn$ (in case of negative answer). The order of queries is crucial and affects $q$.
%\textbf{Given} the set of diagnoses $\mD$ w.r.t. $\langle\mo,\mb,\Tp,\Tn\rangle_\RQ$ and an oracle, \textbf{find} a minimal sequence of queries $(X_1,\dots,X_q)$ to an oracle such that the set of minimal diagnoses $\mD$ can be reduced to the target diagnosis $\dt \in \mD$. The order of queries is crucial and affects the size of the sequence $q$. 
%Given a set of diagnoses $\mD$, a set of logical descriptions $X_j$ is called a \emph{query} iff it is entailed by each ontology $\mo_j$ where $\mo_j \in \setof{\mo \setminus \md_i\,|\,\md_i \in \mD}$.
%Given a set of diagnoses $\mD' \subset \mD$, a set of logical descriptions $X_j$ is called a \emph{query} iff it is entailed by each ontology in $\setof{\mo^{*}_i\,|\,\md_i \in \mD'}$ where $\mo^{*}_i := \mo \setminus \md_i \cup \mb \cup \bigcup_{\tp\in\Tp} \tp$.  
%A set of logical descriptions $X_j$ is called a \emph{query} iff there exists a set of diagnoses $\mD' \subset \mD$ such that $X_j$ is entailed by each ontology in $\setof{\mo^{*}_i\,|\,\md_i \in \mD'}$ where $\mo^{*}_i := \mo \setminus \md_i \cup \mb \cup \bigcup_{\tp\in\Tp} \tp$.  
%
A set of queries for a given set of diagnoses $\mD$ can be generated as shown in Algorithm~\ref{algo_query_gen}. In each iteration, for a set of diagnoses $\dx{} \subset \mD$, the generator gets a set of logical descriptions $X$ that are entailed by each ontology $\mo^{*}_i$ where $\md_i \in \dx{}$ (function \textsc{getEntailments}$)$. These descriptions $X$ are then used to classify the remaining diagnoses in $\mD\setminus\dx{}$ in order to obtain the partition $\langle\dx{},\dnx{},\dz{}\rangle$ associated with~$X$. Then, together with its partition, $X$ is added to the set of queries $\mX$.
% resulting set of queries $\mX = \setof{\tuple{X_j, \dx{j},\dnx{j},\dz{j}}}$ extended with corresponding partitions. 
Note that in real-world applications, investigation of all possible subsets of the set $\mD$ might be infeasible. 
Thus, it is common to approximate the set of all minimal diagnoses by a set of \emph{leading diagnoses}. This set comprises a predefined number $n$ of minimal diagnoses.
%cardinality instead of the set of all minimal diagnoses. 
%is often approximated by a set of \emph{leading diagnoses} 
%limited by a predefined constant $n$.
%of predefined cardinality $n$. 
\begin{algorithm}[tb]
\scriptsize
\KwIn{diagnosis problem instance $\tuple{\mo,\mb,\Tp,\Tn}$, set of diagnoses $\mD$}
\KwOut{a set of queries and associated partitions $\mX$} 
\SetKwFunction{getEnts}{getEntailments}
\ForEach{$\dx{} \subset \mD$}{
      $X \leftarrow \getEnts(\mo, \mb, \Tp, \dx{})$\;
			%Call $R_E$ to get all common entailments $X$ of $\mo \setminus \md_i$ for all $\md_i \in \dx{}$
			\If {$X \neq \emptyset$} {
				\ForEach {$\md_r \in \mD\setminus\dx{}$}{ 
					\lIf { $\mo^{*}_r \,\models X$}{$\dx{} \leftarrow \dx{} \cup \left\{\md_r\right\}$}\;
					\lElseIf	{$\mo^{*}_r \cup X$ is inconsistent}  {$\dnx{} \leftarrow \dnx{} \cup \left\{\md_r\right\}$}\;
					\lElse {$\dz{} \leftarrow \dz{} \cup \left\{\md_r\right\}$}\;
				} %end for each
			$\mX \leftarrow \mX \cup \tuple{X, \dx{}, \dnx{}, \dz{}}$
    } % end if
}
\Return $\mX$\;
\caption{\small Query Generation \normalsize} \label{algo_query_gen}
\normalsize
\end{algorithm}
\begin{algorithm}[b]
\scriptsize
\KwIn{diagnosis problem instance $\langle\mo,\mb,\Tp,\Tn\rangle$, set of diagnoses $\mD$, set of prior fault probabilities $DP$}
\KwOut{target diagnosis $\setof{\dt}$} 
\SetKwFunction{getBestQu}{getBestQuery}
\SetKwFunction{getQuAnswer}{getAnswer}

\Repeat{$|\mD| = 1$}{
      $X \leftarrow$ \getBestQu($\mD, DP$) \;
			%\pose \textbf{'best'} query $X$ to the oracle\;
			\lIf{\getQuAnswer($X$) =\textit{yes}}{
                $\mD \leftarrow \mD \setminus \dnx{}$; $\Tp \leftarrow \Tp \cup \setof{X}$ \;}
			\lElse{$\mD \leftarrow \mD \setminus \dx{}$; $\Tn \leftarrow \Tn \cup \setof{X}$\;}
}
\Return $\mD$\;
\caption{\small Generic Diagnosis Discrimination \normalsize} \label{algo_diag_discrimination}
\normalsize
\end{algorithm}

The query generation algorithm returns a set of queries $\mX$ that generally contains a lot of elements. Therefore the authors in~\cite{jws12} suggested two query selection strategies.

\noindent\textbf{Split-in-half strategy}, selects the query $X_j$ which minimizes the following scoring function: 
\begin{equation*}
sc_{split}(X_j) = \left| |\dx{j}| - |\dnx{j}| \right| + |\dz{j}|
\end{equation*}
I.e. this strategy prefers queries which eliminate half of the diagnoses independently of the query outcome.
%
%The \textbf{entropy-based approach} uses meta information about probabilities $p_t$ that the user makes a fault when using a syntactical construct of type $t \in \mathit{CT}$ where $\mathit{CT}$ is the set of construct types available in the used ontology expression language. For example, $\forall$, $\exists$, $\sqsubseteq$, $\neg$, $\sqcup$, $\sqcap$ are some OWL DL construct types.

\noindent\textbf{Entropy-based strategy}, uses information about prior probabilities $p_t$ for the user to make a fault when using a syntactical construct of type $t \in \mathit{CT}$ where $\mathit{CT}$ is the set of construct types available in the used logical description language. E.g., $\forall$, $\exists$, $\sqsubseteq$, $\neg$, $\sqcup$, $\sqcap$ are some OWL DL construct types. These fault probabilities $p_t$ are assumed to be independent and used to calculate fault probabilities of axioms $ax_k$ as 
%\begin{equation}
%\label{eq:prob_axiom}
$p(\ax_k) = 1 - \prod_{t \in \mathit{CT}} (1-p_t)^{n(t)}$
%\end{equation}
 where $n(t)$ is the number of occurrences of construct type $t$ in $ax_k$.
%The same approach can also be applied in ontology matching. However, in many application scenarios the matched ontologies are assumed to be correct whereas the alignment $M$ is considered as faulty. In this case one can assign some small prior fault probability, e.g. $p(\tax_i)=0.001$, to all axioms $\tax_i$ of the matched ontologies. The prior fault probability of an axiom $\tax_j \in M$ can be computed as $p(\tax_j) = 1-v_j$, where $v_j$ is the confidence value of the correspondence underlying $\tax_j$. 
The probabilities of axioms can in turn be used to determine fault probabilities of diagnoses $\md_i \in \mD$ as 
\vspace{-3pt}
\begin{equation}
\label{eq:prob_diagnosis}
p(\md_i) = \prod_{\ax_r \in \md_i} p(\ax_r) \prod_{\ax_s \in \mo\setminus\md_i} (1-p(\ax_s)).
\end{equation}
\vspace{-9pt}\\
The strategy is then to select the query which minimizes the expected entropy of the set of leading diagnoses $\mD$ after the query is answered. This means that the expected uncertainty is minimized and the expected information gain is maximized. According to \cite{dekleer1987}, this is equivalent to choosing the query $X_j$ which minimizes the following scoring function:
\vspace{0pt}
\begin{equation*} 
    sc_{ent}(X_j) = \sum_{a_j\in\{\textit{yes},\textit{no}\}} p(a_j)\log_2{p(a_j)} + p(\dz{j}) + 1
%\label{eq:score}
\end{equation*} 
\vspace{-5pt}\\
This function is minimized by queries $X_j$ with $p(\dx{j}) = p(\dnx{j}) = 0.5$. So, entropy-based query selection favors queries whose outcome is most uncertain. After each query $X_j$, the diagnosis probabilities are updated according to the Bayesian formula:
\begin{equation} 
    p(\md_i|a_j) = \frac{p(a_j|\md_i)\,p(\md_i)}{p(a_j)} 
\label{eq:bayes}
\end{equation}
\vspace{-5pt}
where 
$a_j \in \{\textit{yes},\textit{no}\}$, 
\begin{equation*}
    p(a_j=\textit{yes}) = \sum_{\md_r \in \dx{j}} p(\md_r) + \frac{1}{2} \sum_{\md_k \in \dz{j}} \md_k
\end{equation*}
and $p(a_j|\md_k):=1/2$ for $\md_k \in \dz{j}$,  $p(a_j|\md_k)=0$ if $\md_k$ is rejected by the query answer $a_j$ and $1$ otherwise.

A generic diagnosis discrimination algorithm (see Algorithm~\ref{algo_diag_discrimination}) can use either of the strategies to identify the target diagnosis $\md_t$. The selection strategy implemented in the \textsc{getBestQuery} function determines the sequence of queries. The result of the evaluation in~\cite{jws12} shows that entropy-based query selection reveals better performance than split-in-half in most of the cases. However, split-in-half proved to be the best strategy in situations when only vague priors are provided, i.e. the target diagnosis $\dt$ has rather low prior fault probability. Therefore selection of prior fault probabilities is crucial for successful query selection and minimization of user interaction.

\noindent\textbf{Example (continued):} 
%To illustrate this, let the user decide to debug our example ontology $\mo$. To that end, the user specifies the two merged ontologies as correct  $p(\tax_i) = 0.001$ for all $\tax_i \in \mo_1\cup\mo_2$ and uses confidence values provided by a matching system to compute the probabilities $p(\tax_5)=0.1$ and $p(\tax_6)=0.15$. Assume that $\md_2$ corresponds to the target diagnosis $\dt$, i.e. the settings provided by the user are incorrect.
To illustrate this, let a user who wants to debug our example ontology $\mo$ 
%regard the two matched ontologies $\mo_1,\mo_2$ as correct and thus 
set $p(\tax_i) := 0.001$ for $ax_{i(i=1,\dots,4)}$ and $p(\tax_5):=0.1, p(\tax_6):=0.15$, e.g. because the user doubts the correctness of $\tax_5,\tax_6$ while being quite sure that $ax_{i(i=1,\dots,4)}$ are correct.
%$\tax_i \in \mo_1\cup\mo_2$ and for the alignment axioms $p(\tax_5):=0.1, p(\tax_6):=0.15$ according to the confidence values provided by the matching system. 
Assume that $\md_2$ corresponds to the target diagnosis $\dt$, i.e. the settings provided by the user are inept.
Application of entropy-based query selection starts with computation of prior fault probabilities of diagnoses\label{ex:probs} $p(\md_1) = p(\md_2) = p(\md_3) = p(\md_4) = 0.003$, $p(\md_5) = 0.393$, $p(\md_6) = 0.591$ (Formula~\ref{eq:prob_diagnosis}). Then $X_1$, i.e. $(\ot \models \{DeptEmployee(s),Student(s)\}?)$, will be identified as the optimal query since it has the minimal score $sc_{ent}(X_1)=0.02$ (see Table~\ref{tab:queries_ex}). 
%However, note that this query could eliminate only two diagnoses $\md_4$ and $\md_6$, if the unfavorable answer \textit{no} is given. Hence, we call $X_1$ a \emph{high-risk query} as its elimination rate = 2/6.
However, since the unfavorable answer $a_1 = \textit{no}$ is given, this query eliminates only two diagnoses $\md_4$ and $\md_6$ (worst case elimination rate $e_{wc}(X_1) = \frac{2}{6}$). 
%Hence, we call $X_1$ a \emph{high-risk query} as its elimination rate = 2/6. 
%
The probability update given by Formula~\ref{eq:bayes} then yields $p(\md_2) = p(\md_3) = p(\md_4) = 0.01$ and $p(\md_5) = 0.97$. As the next query $X_2$ with $sc_{ent}(X_2)=0.811$ is selected and answered unfavorably ($a_2 = \textit{yes}$) as well which results in the elimination of only one single diagnosis $\md_5$ ($e_{wc}(X_2) = \frac{1}{4}$). Since the worst case elimination rate $e_{wc}(X_2)$ is minimal, we call $X_2$ a \emph{high-risk query}. %because the minimal elimination rate is only $\frac{1}{4}$. 
By querying $X_3$ ($sc_{ent}(X_3)=0.082$, $a_3 = \textit{yes}$) and $X_4$ ($sc(X_4)=0$, $a_4 = \textit{yes}$), the further execution of this procedure finally leads to the target diagnosis $\md_2$. So, by applying $sc_{ent}$, four queries are required in order to find $\dt$.
If queries are selected by $sc_{split}$, on the contrary, only three queries are required. The algorithm can select one of the two queries $X_5$ or $X_9$ because each eliminates half of all diagnoses in any case (). We call such a query a \emph{no-risk query}. Let the strategy select $X_5$ which is answered positively ($a_5 = \textit{yes}$). As successive queries, $X_6$ ($a_6=\textit{no}$) and $X_1$ ($a_1=\textit{no}$) are selected, which leads to the revelation of $\dt = \md_2$.

This example demonstrates that the no-risk strategy $sc_{split}$ (\textit{three queries}) is more suitable than $sc_{ent}$ (\textit{four queries}) for fault probabilities which disfavor the target diagnosis. Let us suppose, on the other hand, that probabilities are assigned more reasonably in our example, e.g. $\dt=\md_6$. Then it will take the entropy-based strategy only \textit{two queries} $(X_1,X_6)$ to find $\dt$ while split-in-half will still require \textit{three queries}, e.g. $(X_5,X_1,X_6)$. The complexity of $sc_{ent}$ in terms of required queries varies between $O(1)$ in the best and $O(|\mD|)$ in the worst case depending on the appropriateness of the fault probabilities. In contrast, $sc_{split}$ always requires $O(\log_2 |\mD|)$ queries.

We learn from this example that the best choice of discrimination strategy depends on the quality of the meta information in terms of prior fault probabilities. 
%In case of an ontology elaborated by a single developer, logs about user faults, for example, could serve as a guidance for specifying the priors.
%
In cases where adequate meta information is not available and hard to estimate, e.g. ontology alignment and ontology learning, the inappropriate choice of strategy might cause tremendous extra effort for the user interacting with the debugging system.
%However, in the case of  ontology alignment or collaborative ontology development, where standalone ontologies are developed, maintained and matched by different users, estimation of the correct prior fault probabilities might be problematic. 
Therefore, we suggest to exploit additional information gathered by querying the oracle in order to estimate the quality of given meta information.
The new strategy we present incorporates the elimination rate achieved by the current query when choosing the successive query. To this end, a parameter of maximum allowed \emph{query-risk} is permanently adapted. Our method combines the advantages of both the entropy-based approach and the split-in-half approach. On the one hand, it exploits the given prior fault probabilities if they are of high quality. On the other hand, it quickly loses trust in the priors and gets more cautious if some evidence is given that the probabilities are misleading. 
%Thus, the approach is perfectly suited for the use case of debugging aligned ontologies.
%\vspace{-10pt}
\begin{table}
\centering 
\begin{tabular}{@{\extracolsep{0pt}}l|l|l|c}
Query & $\dx{i}$  & $\dnx{i}$ &  $\dz{i}$ \\\hline
$X_1:\{DeptEmployee(s),$ & $\md_4,\md_6$ & $\md_1,\md_2,\md_3,\md_5$ & $\emptyset$ \\
$\qquad\quad Student(s)\}$ & & & \\
$X_2:\{PhD(s)\}$ & $\md_1, \md_2,\md_3,\md_4,\md_6$   & $\md_5$ & $\emptyset$ \\
$X_3:\{Researcher(s)\}$ & $\md_2, \md_3, \md_4, \md_6$  & $\md_1, \md_5$ & $\emptyset$ \\
$X_4:\{Student(s)\}$ & $\md_1, \md_2, \md_4, \md_5, \md_6$ & $\md_3$ & $\emptyset$ \\
$X_5:\{Researcher(s),$ & $\md_2, \md_4, \md_6$ & $ \md_1, \md_3, \md_5$ &
$\emptyset$ \\
$\qquad\quad Student(s)\}$ & & & \\
$X_6:\{DeptMember(s)\}$ & $\md_3, \md_4$ & $ \md_1, \md_2, \md_5, \md_6$ & \\
$X_7:\{PhD(s),$ & $\md_1, \md_2, \md_4, \md_6$ & $ \md_3, \md_5$ &
$\emptyset$ \\
$\qquad\quad Student(s)\}$ & & & \\
$X_8:\{DeptMember(s),$ & $\md_2$ & $ \md_1, \md_3, \md_4, \md_5, \md_6$ &
$\emptyset$ \\
$\qquad\quad Student(s)\}$ & & & \\
$X_9:\{DeptEmployee(s)\}$ & $\md_3, \md_4, \md_6$ & $ \md_1, \md_2, \md_5$ &
$\emptyset$ \\\hline
\end{tabular} \vspace{2pt}
\caption{Nine queries computed with respect to entailed assertional axioms for diagnoses $\md_i \in \mD$ of the sample ontology $\mo$. Given that no diagnoses have been eliminated yet, $X_2, X_4, X_8$ are high-risk queries, $X_5, X_9$ are no-risk queries.}\label{tab:queries_ex}
%\vspace{-20pt}
\end{table}
\vspace{-11pt}

%% file: strategy.tex
\vspace{-10pt}
\section{Risk Optimization Strategy for Query Selection}
\label{sec:theory} 
The proposed Risk Optimization Algorithm (RIO) extends entropy-based query selection strategy with a dynamic learning procedure that learns by reinforcement how to select optimal queries. Moreover, it continually improves the prior fault probabilities based on new knowledge obtained through queries to a user.
The behavior of our algorithm can be co-determined by the user. The algorithm takes into account the user's doubt about the priors in terms of the initial cautiousness $\uc$ as well as the cautiousness interval $[\underline{\uc},\overline{\uc}]$ where $\uc,\underline{\uc},\overline{\uc}\in [\uc_{\min},\uc_{\max}]:=[0, \left\lfloor |\mD|/2\right\rfloor/|\mD|]$, $\underline{\uc}\leq\uc\leq\overline{\uc}$ and $\mD$ contains at most $n$ leading diagnoses (see Section~\ref{sec:basics}). The interval $[\underline{\uc},\overline{\uc}]$ constitutes the set of all admissible cautiousness values the algorithm may take during the debugging session. High trust in the prior fault probabilities is reflected by specifying a low minimum required cautiousness $\underline{\uc}$
and/or a low maximum admissible cautiousness $\overline{\uc}$. 
If the user is unsure about the rationality of the priors this can be expressed by setting $\underline{\uc}$ and/or $\overline{\uc}$ to a higher value. 
%By contrast, the value assigned to $\overline{\uc}$ indicates the maximum admissible cautiousness of the algorithm. Low values of $\overline{c}$ can be interpreted as trust in the priors as it forces the algorithm to avoid low-risk queries. 
Intuitively, $\underline{\uc} - \uc_{\min}$ and $\uc_{\max} - \overline{\uc}$ represent the minimal desired difference in performance to a high-risk (entropy) and no-risk (split-in-half) query selection, respectively. 

The relationship between cautiousness $\uc$ and queries is formalized by the following definitions:
\begin{definition}[Cautiousness of a Query]\label{def:query_cautiousness}
We define the \emph{cautiousness} $\qc(X_i)$ of a query $X_i$ as follows:
\vspace{-9pt}
\begin{equation*}\label{eq:query_cautiousness}
\qc(X_i) := \frac{ \text{min}\left\{|\dx{i}| ,|\dnx{i}| \right\} }{|\mD|} \in \left[0,\frac{\left\lfloor \frac{|\mD|}{2}\right\rfloor}{|\mD|}\right]
\end{equation*}
A query $X_i$ is called \emph{braver} than query $X_j$ iff $\qc(X_i) < \qc(X_j)$. Otherwise $X_i$ is called \emph{more cautious} than $X_j$.
A query with highest possible cautiousness is called \emph{no-risk query}.
\end{definition}
\begin{definition}[Elimination Rate]\label{def:elimination_rate}
Given a query $X_i$ and the corresponding answer $a_i \in \{\textit{yes},\textit{no}\}$, the \emph{elimination rate} $e(X_i,a_i)$ is defined as follows:
\[
e(X_i,a_i)=\begin{cases}
  \frac{|\dnx{i}|}{|\mD|}  	& \text{if }\;a_i = \textit{yes}   \vspace{0.2cm}\\ 
  \frac{|\dx{i}|}{|\mD|} 		& \text{if }\;a_i = \textit{no}
\end{cases}
\]
The answer $a_i$ to a query $X_i$ is called \emph{favorable} iff it maximizes the elimination rate $e(X_i,a_i)$. Otherwise $a_i$ is called \emph{unfavorable}. The minimal or worst case elimination rate $\min_{a_i \in \{\textit{yes},\textit{no}\}}(e(X_i,a_i))$ of $X_i$ is denoted by $e_{wc}(X_i)$.
\end{definition}
So, the cautiousness $\qc(X_i)$ of a query $X_i$ is exactly the minimal, i.e.~worst case, elimination rate, i.e. $\qc(X_i) = e_{wc}(X_i) = e(X_i,a_i)$ given that $a_i$ is the unfavorable query result.
Intuitively, the user-defined cautiousness $\uc$ is the minimum proportion of diagnoses in $\mD$ which should be eliminated by the successive query.
For braver queries the interval between minimum and maximum elimination rate is 
larger than for more cautious queries. For no-risk queries it is minimal. 
\begin{definition}[High-Risk Query]\label{def:high-risk_query}
Given a query $X_i$ and cautiousness $\uc$, then $X_i$ is called a \emph{high-risk query} iff $\qc(X_i) < \uc$, i.e. the cautiousness of the query is lower than the algorithm's current cautiousness value $c$.
Otherwise, $X_i$ is called \emph{non-high-risk query}. By $\HR_c(\mX)\subseteq \mX$ we denote the set of all high-risk queries w.r.t. $c$. For given cautiousness $c$, the set of all queries $\mX$ can be partitioned in high-risk queries and non-high-risk queries.
\end{definition}

\label{ex:high-risk_query}
\noindent\textbf{Example (continued):} Reconsider the example given in Section~\ref{sec:basics}. Let the user specify $\uc = 0.3$ for the set $\mD$ including $n=6$ diagnoses. Given these settings, $X_1$ is a non-high-risk query since its cautiousness $\qc(X_1) = 2/6 \geq 0.3 = \uc$. The query $X_8$ is a high-risk query because $\qc(X_2) = 1/6 < 0.3 = \uc$ and $X_5$ is a no-risk query due to $\qc(X_5) = 3/6 = \lfloor\frac{|\mD|}{2}\rfloor/|\mD|$.

Given a user's answer $a_s$ to a query $X_s$, the cautiousness $\uc$ is updated depending on the elimination rate $e(X_s,a_s)$ by $\uc \leftarrow \uc + \uc_{adj}$
%as follows:
%
%\begin{equation}\label{eq:risk_update}
	%\uc \;\leftarrow\; \uc + \uc_{adj}
%\end{equation}
%
where $\uc_{adj}$ denotes the cautiousness adjustment factor which is defined as follows:
\begin{equation}
	\uc_{adj} :=\; 2\,(\overline{\uc} - \underline{\uc}) \mathit{adj} 	
\label{eq:adjust}
\end{equation}
The factor $2 \, (\overline{\uc} - \underline{\uc})$ in Formula \ref{eq:adjust} is a scaling factor that simply regulates the extent of the cautiousness adjustment depending on the interval length $\overline{\uc} - \underline{\uc}$. The more crucial factor in the formula is $\mathit{adj}$ which indicates the sign and magnitude of the cautiousness adjustment.  
\begin{equation*}
\mathit{adj} :=  \frac{\left\lfloor \frac{|\mD|}{2}-\epsilon\right\rfloor}{|\mD|} - e(X_s,a_s)
\end{equation*}
where $\epsilon \in (0,\frac{1}{2})$ is a constant which prevents the algorithm from getting stuck in a no-risk strategy for even $|\mD|$. E.g., given $\uc=0.5$ and $\epsilon=0$, the elimination rate of a no-risk query $e(X_s,a_s) = \frac{1}{2}$ resulting always in $adj=0$. 
The value of $\epsilon$ can be set to an arbitrary real number, e.g. $\epsilon := \frac{1}{4}$. 
%
%Note that $\mathit{adj} < 0$ holds for any favorable query result, independently of the cautiousness $\qc(X_s)$ of the asked query $X_s$. This has the following reasons: For even $|\mD|$ and any favorable query outcome $a_s$, the resulting elimination rate $e(X_s,a_s) \geq \lfloor \frac{|\mD|}{2}\rfloor / |\mD|$, which is the maximal value the minimal elimination rate of a query can take (see Definition \ref{def:query_cautiousness}). So, $\lfloor \frac{|\mD|}{2} - \epsilon\rfloor / |\mD| < \lfloor \frac{|\mD|}{2}\rfloor / |\mD| \leq e(X_s,a_s)$ holds since $|\mD|$ is even and $\epsilon \in (0,\frac{1}{2})$. In case of odd $|\mD|$, for any favorable query result $a_s$: $e(X_s,a_s) > \lfloor \frac{|\mD|}{2}\rfloor / |\mD| = \lfloor \frac{|\mD|}{2} - \epsilon\rfloor / |\mD|$. 
%
If $\uc + \uc_{adj}$
%the value $\uc$ computed in (\ref{eq:risk_update}) 
is outside the user-defined cautiousness interval $[\underline{\uc},\overline{\uc}]$, it is set to $\underline{\uc}$ if $\uc < \underline{\uc}$ and to $\overline{\uc}$ if $\uc > \overline{\uc}$. Positive $\uc_{adj}$ is a penalty telling the algorithm to get more cautious, whereas negative $\uc_{adj}$ is a bonus resulting in a braver behavior of the algorithm.

\noindent\textbf{Example (continued):} Assume that an expert is quite unsure about the location of the fault and thus
%has doubts whether the fault is in the alignment and 
sets $\uc=0.4$, $\underline{\uc}=0$ and $\overline{\uc}=0.5$. In this case the algorithm selects a no-risk query $X_5$ just as the split-in-half strategy. Given 
%the answer 
$a_5=\textit{yes}$ and $|\mD|=6$, the algorithm computes the elimination rate $e(X_5,\textit{yes})=0.5$ and adjusts the cautiousness by $\uc_{adj}=-0.17$ which yields $\uc=0.23$. This allows RIO to select a higher-risk query in the next iteration. The algorithm finds the target diagnosis $\dt=\md_2$ by asking three queries.

\begin{algorithm}[b]
\scriptsize
\KwIn{diagnosis problem instance $\langle\mo,\mb,\Tp,\Tn\rangle$, fault probabilities of diagnoses $DP$, cautiousness $C=(\uc,\underline{\uc},\overline{\uc})$, number of leading diagnoses $n$ to be considered, acceptance threshold $\sigma$
}
\KwOut{a diagnosis $\md$} 

\SetKwFunction{getMinScQ}{getMinScoreQuery}
\SetKwFunction{getScore}{eliminationRate}
\SetKwFunction{aboveThresh}{aboveThreshold}
\SetKwFunction{mostProbDiag}{mostProbableDiag}
\SetKwFunction{getPercent}{getQueryCautiousness}
\SetKwFunction{getAlpha}{getMinAlpha}
\SetKwFunction{getBestQ}{getAlternativeQuery}
\SetKwFunction{performAdapt}{performAdaptation}
\SetKwFunction{getAnswer}{getAnswer}
\SetKwFunction{updateProbs}{updateProbablities}
\SetKwFunction{updateRisk}{updateCautiousness}
\SetKwFunction{getDiagnoses}{getDiagnoses}
\SetKwFunction{genQs}{generateQueries}
\SetKwFunction{compPriors}{getProbabilities}
$\Tp \leftarrow \emptyset$;
$\Tn \leftarrow \emptyset$;
$\mD \leftarrow \emptyset$\;
\Repeat{$(\aboveThresh(DP,\sigma) \lor \getScore(X_s) = 0$)}{
        $\mD \leftarrow \getDiagnoses(\mD, n, \mo, \mb, \Tp, \Tn)$\;
				$DP \leftarrow \compPriors(DP,\mD, \Tp, \Tn)$\;
				$\mX \leftarrow \genQs(\mo, \mb, \Tp, \mD)$\;
				$X_s \leftarrow \getMinScQ(DP,\mX)$\;    
        %$s\leftarrow\getScore(X_s,sc_{ent})$\;
				\lIf {$\getPercent(X_s,\mD) < \uc$}{        
						$X_s \leftarrow \getBestQ(\uc,\mX, DP, \mD)$\;
				}  
				\lIf {$ \getAnswer(X_s) = \textit{yes}$}{        
						$\Tp \leftarrow \Tp \cup \{X_s\}$\;  
				}
				\lElse{
						$\Tn \leftarrow \Tn \cup \{X_s\}$\;
				} 
				$c \leftarrow \updateRisk(\mD,\Tp,\Tn,X_s,\uc,\underline{\uc},\overline{\uc})$\;
}
\Return $\mostProbDiag(\mD, DP)$\;
\caption{Risk Optimization Algorithm (RIO)} \label{algo_main}
\normalsize
\end{algorithm}

The RIO algorithm, described in Algorithm \ref{algo_main}, starts with the computation of minimal diagnoses. \textsc{getDiagnoses} function implements a combination of hitting-set (HS-Tree)~\cite{Reiter87} and QuickXPlain~\cite{junker04} algorithms as suggested in~\cite{jws12}. Using uniform cost search, the algorithm extends the set of leading diagnoses $\mD$ with a maximum number of most probable minimal diagnoses such that $|\mD| \leq n$. 

Then the \textsc{getProbabilities} function calculates the fault probabilities $p(\md_i)$ for each diagnosis $\md_i$ of the set of leading diagnoses $\mD$ using Formula~\ref{eq:prob_diagnosis}. In order to take into account all information gathered by querying an oracle so far the algorithm adjusts fault probabilities $p(\md_i)$ as follows: $p_{adj}(\md_i)=(1/2)^{z}\, p(\md_i)$, where $z$ is the number of precedent queries $X_k$ for which $\md_i \in \dz{k}$.
Afterwards the probabilities $p_{adj}(\md_i)$ are normalized. Note that $z$ can be computed from $\Tp$ and $\Tn$ which comprise all query answers. This way of updating probabilities is exactly in compliance with the Bayesian theorem given by Formula~\ref{eq:bayes}.
Based on the set of leading diagnoses $\mD$, \textsc{generateQueries} generates all queries according to Algorithm \ref{algo_query_gen}.
	\textsc{getMinScoreQuery} determines the best query $\Xsc \in \mX$ according to $sc_{ent}$. That is:
			\begin{equation*}
				\Xsc = \argmin_{X_k \in \mX}(sc_{ent}(X_k)) 
			\end{equation*}	
	\label{step_standard} 
If $\Xsc$ is a non-high-risk query, i.e. $\uc \leq \qc(\Xsc)$ (determined by \textsc{getQueryCautiousness}), $\Xsc$ is selected. In this case, $\Xsc$ is the query with maximum information gain among all queries $\mX$ and additionally guarantees the required elimination rate specified by $\uc$.
	
\label{step_alternative} Otherwise, \textsc{getAlternativeQuery} selects the query $\Xalt \in \mX$\; $(\Xalt \neq \Xsc)$ which has minimal score $sc_{ent}$ among all least cautious non-high-risk queries $L_c$. I.e.:
\begin{equation*}
	\Xalt = \argmin_{X_k \in \mathit{L}_c}(sc_{ent}(X_k)) 
\end{equation*}
where $\mathit{L_c} = \{X_r \in \mX \setminus \HR_c(\mX) \;|\; \forall X_t \in \mX \setminus \HR_c(\mX):\, \qc(X_r) \leq \qc(X_t)\}$. If there is no such query $\Xalt\in\mX$, then $\Xsc$ is selected.
	
Given the positive answer of the oracle, the selected query $X_s \in \setof{\Xsc,\Xalt}$ is added to the set of positive test cases $\Tp$ or, otherwise, to the set of negative test cases $\Tn$. In the last step of the main loop the algorithm updates the cautiousness value $\uc$ (function \textsc{updateCautiousness}) as described above.

Before the next query selection iteration starts, a stop condition test is performed. The algorithm evaluates whether the most probable diagnosis is at least $\sigma\%$ more likely than the second most probable diagnosis (\textsc{aboveThreshold}) or none of the leading diagnoses has been eliminated by the previous query, i.e.\textsc{getEliminationRate} returns zero for $X_s$. In case that one of the stop conditions is fulfilled, the presently most likely diagnosis is returned (\textsc{mostProbableDiag}).

%% file: eval.tex
\section{Evaluation} \label{sec:eval}
The main points we want to show in this evaluation are: On the one hand, independently of the specified meta information, RIO 
%in most cases manifests a performance as good as or better than the best strategy of entropy-based method and split-in-half 
exhibits superior average behavior compared to entropy-based method and split-in-half w.r.t. the amount of user interaction required. On the other hand, we want to demonstrate that RIO scales well and that the reaction time measured is well suited for an interactive debugging approach.

As data source for the evaluation we used problematic real-world ontologies produced by ontology matching systems.\footnote{Thanks to Christian Meilicke for the supply of the test cases used in the evaluation.} This has the following reasons: (1) Matching results often cause inconsistency and/or incoherency of ontologies. 
%(2) which are often of large size and 
(2) The (fault) structure of different ontologies obtained through matching generally varies due to different authors and matching systems involved in the genesis of these ontologies. (3) For the same reasons, it is hard to estimate the quality of fault probabilities, i.e. it is unclear which of the existing query selection strategies to chose for best performance. (4) Available reference mappings can be used as correct solutions of the debugging procedure.

Note that the comparison of RIO with techniques integrated in ontology matching systems such as CODI~\cite{noessner2010} or LogMap~\cite{Jimenez-Ruiz2011} is inappropriate, since all these systems use greedy diagnosis techniques (e.g.~\cite{MeilickeStuck2009}), whereas the method presented in this paper is complete.

%%%%%%%%%%%%%%%%%%%%%%%%%%%%%%%%
Matching of two ontologies $\mo_i$ and $\mo_j$ is usually understood as detection of correspondences between elements of these ontologies~\cite{Shvaiko2012}: 
\begin{definition}[Ontology matching]\label{def:align} Let $Q(\mo_i)$ and $Q(\mo_j)$ denote the sets of matchable elements in ontologies $\mo_i$ and $\mo_j$. An ontology matching operation determines an \emph{alignment} $\Align_{ij}$, which is a set of correspondences between matched ontologies $\mo_i$ and $\mo_j$. Each \emph{correspondence} is a 4-tuple $\tuple{x_i,x_j,r,v}$, such that $x_i \in Q(\mo_i)$, $x_j \in Q(\mo_j)$, $r$ is a semantic relation and $v \in [0,1]$ is a confidence value. We call $\mo_{i\Align j} := \mo_i \cup \Align_{ij} \cup \mo_j$ the \emph{aligned ontology} for $\mo_i$ and $\mo_j$.
\end{definition}
In our approach the elements of $Q(\mo)$ are restricted to atomic concepts and roles and $r \in \setof{\sqsubseteq, \sqsupseteq, \equiv}$ under the natural alignment semantics~\cite{MeilickeStuck2009} that maps correspondences one-to-one to axioms of the form $x_i~r~x_j$. 
%However, as mentioned above, application of modern ontology matching systems can result in faulty ontologies. 

\noindent\textbf{Example (continued):} Imagine that our example ontology $\mo$ evolved from matching two standalone ontologies $\mo_1 := \{\tax_1,\tax_2\}$ and $\mo_2 := \{\tax_3,\tax_4\}$ resulting in the alignment $\Align_{12} = \{\tax_5,\tax_6\}$. As a concrete use case, for instance, assume two departments of a university, each developing an ontology for their homepage where $\mo_1$ is an excerpt of the first ontology and $\mo_2$ an excerpt of the second. In order to unite the homepages and underlying ontologies, an ontology matching system could be consulted. However, if, as in this case, an alignment $\Align_{12}$ is generated which yields an inconsistent aligned ontology $\mo_{1\Align 2}$, the output of the matching system as-is is useless and combining the homepages is impossible without according ontology debugging support. If we recall the set of diagnoses for $\mo$ consisting of all single axioms in $\mo$, we realize that the fault we are trying to find may be located either in $\mo_1$ or in $\mo_2$ or in $\Align_{12}$. Existing approaches to alignment debugging usually consider only the produced alignment as problem source. Our approach, on the contrary, is designed to cope with the most general setting: Any subset $S \subseteq \mo_{1\Align 2}$ of axioms of the aligned ontology can be analyzed for faults whereas $\mo_{1\Align 2}\setminus S$ can be added to the background axioms $\mb$, if known to be correct. In this way, the search space for diagnoses can be restricted elegantly depending on the prior knowledge about $\dt$, which can greatly reduce the complexity of the underlying diagnosis problem.
%In the situation where an expert is sure that $\dt \subseteq \mo' \subset \mo_{ij}$, one can simply add $\mo_{ij}\setminus \mo'$ to the background axioms $\mb$. 
%\footnote{Notation: $x_i$ means that concept/property/individual $x$ belongs to ontology $\mo_i$; mapping axioms include concepts from both ontologies.}
%The obtained diagnoses mean that the cause for the inconsistency of $\mo$ can be located in $\mo_1$, $\mo_2$ or in the mapping $M$.
%%%%%%%%%%%%%%%%%%%%%%%%%%%%%%%%%

In \cite{Stuckenschmidt2008} it was shown that existing debugging approaches suffer from serious problems w.r.t. both scalability and correctness of results when tested on a dataset of incoherent aligned OWL ontologies. Since RIO is an \emph{interactive} ontology debugging approach able to query and incorporate additional information into its computations, it can cope with cases unsolved in \cite{Stuckenschmidt2008}. In order to provide evidence for this and to show the feasibility of RIO -- simultaneously to the main goals of this evaluation -- we decided to use a superset of the dataset\footnote{http://code.google.com/p/rmbd/downloads} used in \cite{Stuckenschmidt2008} for our tests. 
%In order to 
%provide a comparison of RIO to 
%make evident that our approach succeeds where others failed
%other approaches 
%
%Note that, albeit the experiments in ---\cite{Stuckenschmid}--- were conducted in 2008, to the best of our knowledge no better results to the ones achieved in \cite{}
%there have been no publications of enhancements to the tested debugging approaches in the meantime. 
%I.e., the comparison is valid.
%The dataset includes ??????? incoherent OWL ontologies.
%The debugging methods were tested using the OntoFarm dataset\footnote{http://code.google.com/p/rmbd/downloads} comprising six ontologies $Ont = \{\text{CRS}, \text{PCS}, \text{CMT}, \text{CONFTOOL}, \text{SIGKDD}, \text{EKAW}\}$ in the domain of conference organization.
%For the evaluation of RIO we made experiments based on a test dataset\footnote{http://code.google.com/p/rmbd/downloads} 
%which comprises 26 incoherent OWL ontologies. 
%
Each incoherent aligned ontology $\mo_{i\Align j}$ in the dataset is the result of applying one of the ontology matching systems COMA++, Falcon-AO, HMatch or OWL-CTXmatch to a set of six ontologies $Ont = \{\text{CRS}, \text{PCS}, \text{CMT}, \text{CONFTOOL}, \text{SIGKDD}, \text{EKAW}\}$ in the domain of conference organization. For a given pair of ontologies $\mo_i\neq\mo_j \in Ont$, each system produced an alignment $M_{ij}$.
%
%\footnote{The same datasets and matching tools were used in the evaluation in \cite{Meilicke2008a}.}
%
On the basis of a manually produced reference alignment 
%(used in the evaluation of \cite{Meilicke2008a}) 
$\mathcal{R}_{ij} \subseteq M_{ij}$ for ontologies $\mo_i,\mo_j$ (cf. \cite{Meilicke2008a}), we were able to fix a target diagnosis $\dt$ for each incoherent $\mo_{i\Align j}$. In cases where $\mathcal{R}_{ij}$ suggested a non-minimal diagnosis, we defined $\dt$ as the minimum cardinality diagnosis which was a subset of $M_{ij}\setminus \mathcal{R}_{ij}$. In one single case, $\mathcal{R}_{ij}$ proved to be incoherent because an obviously valid correspondence $Reviewer_1 \equiv reviewer_2$ turned out to be incorrect. We re-evaluated this ontology and specified a coherent $\mathcal{R}_{ij}$. Yet this makes evident that, in general, people are not capable of analyzing alignments without adequate tool support. 

In our experiments we set the prior fault probabilities as follows: $p(ax_k) := 0.001$ for $\tax_k \in \mo_i \cup \mo_j$ and $p(\tax_m):= 1-v_m$ for $\tax_m \in \Align_{ij}$, where $v_m$ is the confidence of the correspondence underlying $\tax_m$. Note that this choice results in a significant bias towards diagnoses which include axioms from $M_{ij}$. Based on these settings, in the first experiment (EXP-1), we simulated an interactive debugging session employing split-in-half (SPL), entropy (ENT) and RIO algorithms, respectively, for each 
%of the 26 
ontology $\mo_{i\Align j}$. Throughout all experiments, we performed module extraction \cite{Grau2008b} before each test run, which is a standard preprocessing method for ontology debugging approaches. All tests were executed on a Core-i7 (3930K) 3.2Ghz, 32GB RAM and with Ubuntu Server 11.04 and Java 6 installed. The number $|\mD|$ of leading diagnoses was set to 9 and $\sigma:=85\%$. As input parameters for RIO we set $\uc:= 0.25$ and $[\underline{c},\overline{c}] := [\uc_{\min},\uc_{\max}] = [0,\frac{4}{9}]$. For the tests we considered the most general setting, i.e. $\dt \subset \mo_{i \Align j}$. So, we did not restrict the search for $\dt$ to $\Align_{ij}$ only, simulating the case where the user has no idea whether any of the input ontologies $\mo_i,\mo_j$ or the alignment $\Align_{ij}$ or a combination thereof is faulty.
%, contrary to \cite{Meilicke2008a}. 
In each test run we measured the number of required queries until $\dt$ was identified. In order to simulate the case where the fault includes at least one axiom $\tax \in \mo_{i\Align j} \setminus M_{ij}$, we implemented a second test session with altered $\dt$. In this experiment (EXP-2), we precalculated a maximum of $30$ most probable minimal diagnoses, and from these we selected the diagnosis with the highest number of axioms $\tax_k \in \mo_{i\Align j} \setminus M_{ij}$ as $\dt$ in order to simulate more unsuitable meta information. All the other settings were left unchanged. The queries generated in the tests were answered by an automatic oracle by means of the target ontology $\mo_{i\Align j}\setminus \dt$. 
%Table \ref{tab:onto_metrics} presents the average values of metrics of the ontologies $\mo_{i\Align j}$ concerning number of axioms, concepts and properties, and number and minimal/maximal size of diagnoses and conflict sets, respectively.
The average metrics for the set of aligned ontologies $\mo_{i\Align j}$ per matching system were as follows: $312 \leq |\mo_{i\Align j}| \leq 377$ and $19.1 \leq |\Align_{ij}| \leq 28.4$.

In order to analyze the scalability of RIO, we used the set of ontologies
%\footnote{The raw data representing the output of matching systems was downloaded from http://bit.ly/Koh1NB. The reference alignment as well as the source ontologies Mouse and Human were downloaded from http://bit.ly/MU5Ca9.} 
from the ANATOMY track in the Ontology Alignment Evaluation Initiative\footnote{http://oaei.ontologymatching.org} (OAEI) 2011.5, which comprises two input ontologies $\mo_1$ (Human, 11545 axioms) and $\mo_2$ (Mouse, 4838 axioms). The size of the alignments generated by 12 different matching systems was between 1147 and 1461 correspondences. Note that the aligned ontologies output by five matching systems, i.e. CODI, CSA, MaasMtch, MapEVO and Aroma, could not be analyzed in the experiments. This was due to a consistent output produced by CODI and the problem that the reasoner was not able to find a model within acceptable time (2 hours) in the case of CSA, MaasMtch, MapEVO and Aroma. Similar reasoning problems were also reported in \cite{Ferrara2011}.
%The ontologies, for which the alignments were calculated, include 11545 (Human) and 4838 (Mouse) axioms, respectively. 
Given the ontologies $\mo_1$ and $\mo_2$, the output $\Align_{12}$ of a matching system, and the correct reference alignment $\mathcal{R}_{12}$, we first fixed $\dt$ as follows: Both ontologies $\mo_1$ and $\mo_2$ as well as the correctly extracted alignments $\Align_{12} \cap \mathcal{R}_{12}$ were placed in the background knowledge $\mb$. The incorrect correspondences $\Align_{12}\setminus \mathcal{R}_{12}$ were analyzed by the debugger. In this way, we identified a set of diagnoses, where each diagnosis is a subset of $\Align_{12}\setminus \mathcal{R}_{12}$. From this set of diagnoses, we randomly selected one diagnosis as $\dt$. 
Then we started the actual experiments:
In EXP-3\footnote{For all details w.r.t. $\langle\text{EXP-3,EXP-4}\rangle$, see http://code.google.com/p/rmbd/wiki/ OntologyAlignmentAnatomy.}, in order to simulate reasonable prior fault probabilities, a debugging session with parameter settings as in EXP-1 was executed. In EXP-4, we altered the settings in that we specified $p(\tax_k) := 0.01$ for $\tax_k \in \mo_i \cup \mo_j$ and $p(\tax_m):= 0.001$ for $\tax_m \in \Align_{ij}$, which caused the target diagnosis, that consisted solely of axioms in $\Align_{ij}$, to get assigned a relatively low prior fault probability. 
%Note that precalculating 30 diagnoses as in EXP-2
%The results are summarized in --- Table --- which shows that the reaction time is still practical.....

\begin{figure}[ht]
\centering
\subfigure[]{
    %\vspace{-25mm}
    \begin{tabular}[b]{@{\extracolsep{0pt}} l|c|c||c|c}
																												& EXP-1		& EXP-2 	& EXP-3			& EXP-4			\\	\hline
			$q_\text{SPL} < q_\text{ENT}$ 										& 11\%	  & 37\% 		& 	0\%			&   29\%		\\
			$q_\text{ENT} < q_\text{SPL}$ 										& 81\%    &	56\% 	  & 	100\%		&		71\%		\\
			$q_\text{SPL} = q_\text{ENT}$											& 7\%     &	7\% 	  & 	0\%			&		0\%			\\
			$q_\text{RIO} < \min$															& 4\%	  	& 26\% 	  & 	29\%		&		71\%		\\
			$q_\text{RIO} \leq \min$															& 74\%		& 74\% 	  & 	100\%		&		100\%		\\	\hline 
				\multicolumn{5}{c}{}	\\
																											
		\end{tabular}
		\vspace{15pt}
		\label{tab:best_strategies}
}
\subfigure[]{
    %\rule{4cm}{3cm}
    \includegraphics[width=0.45\textwidth, trim=2mm 0mm 0mm 15mm]{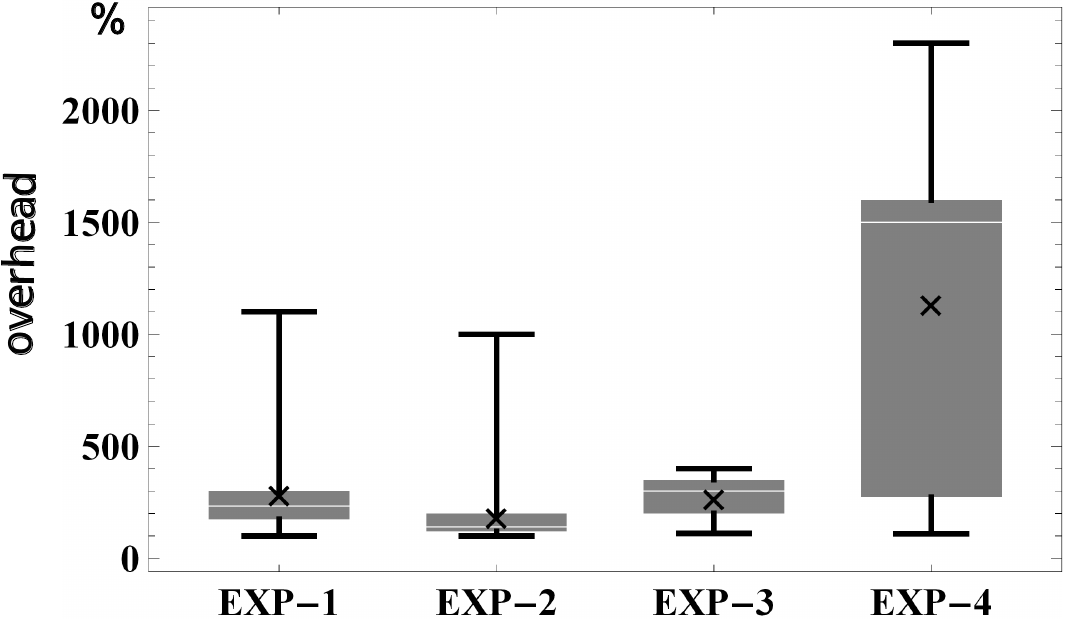}
    \label{fig:diff_SPL_ENT}
}
\vspace{-8pt}
\caption[]{\textbf{\subref{tab:best_strategies}} Percentage rates indicating which strategy performed best/better w.r.t. the required user interaction, i.e. number of queries. EXP-1 and EXP-2 involved 27, EXP-3 and EXP-4 seven debugging sessions each. $q_{str}$ denotes the number of queries needed by strategy $str$ and $\min$ is an abbreviation for $\min(q_\text{SPL},q_\text{ENT})$. \textbf{\subref{fig:diff_SPL_ENT}} Box-Whisker Plots presenting the distribution of overhead $(q_w-q_b)/q_b*100$ (in \%) per debugging session of the worse strategy $q_w := \max(q_\text{SPL},q_\text{ENT})$ compared to the better strategy $q_b := \min(q_\text{SPL},q_\text{ENT})$. Mean values are depicted by a cross.}
\label{fig:best_diff}
\end{figure}

\begin{figure}[ht]
\centering
\subfigure[]{
    %\vspace{-15pt}
    \includegraphics[width=0.46\textwidth,trim=7mm -12mm 1mm 20mm]{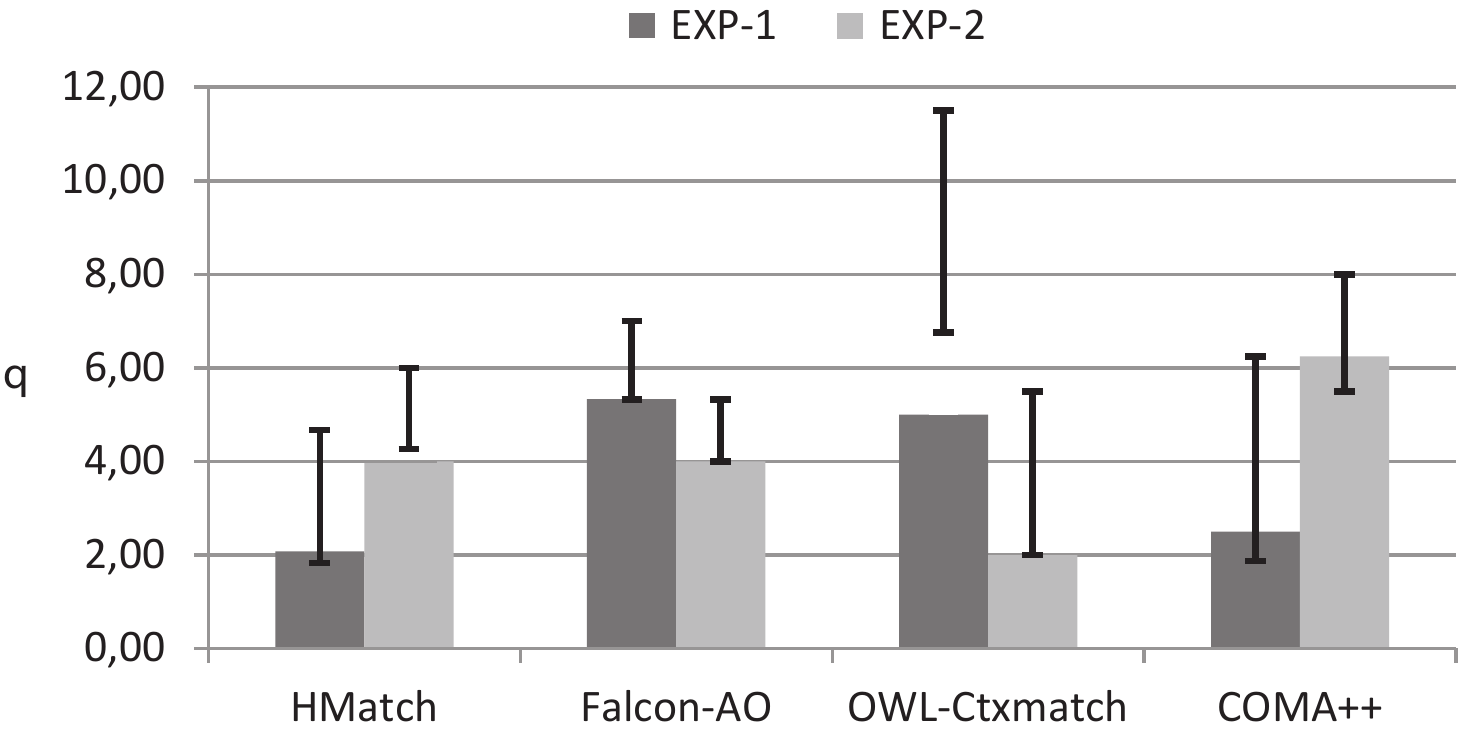}
		\label{fig:query_conference}
}
\subfigure[]{
    %\rule{4cm}{3cm}
    \includegraphics[width=0.46\textwidth,trim=5mm 7mm 1mm 25mm]{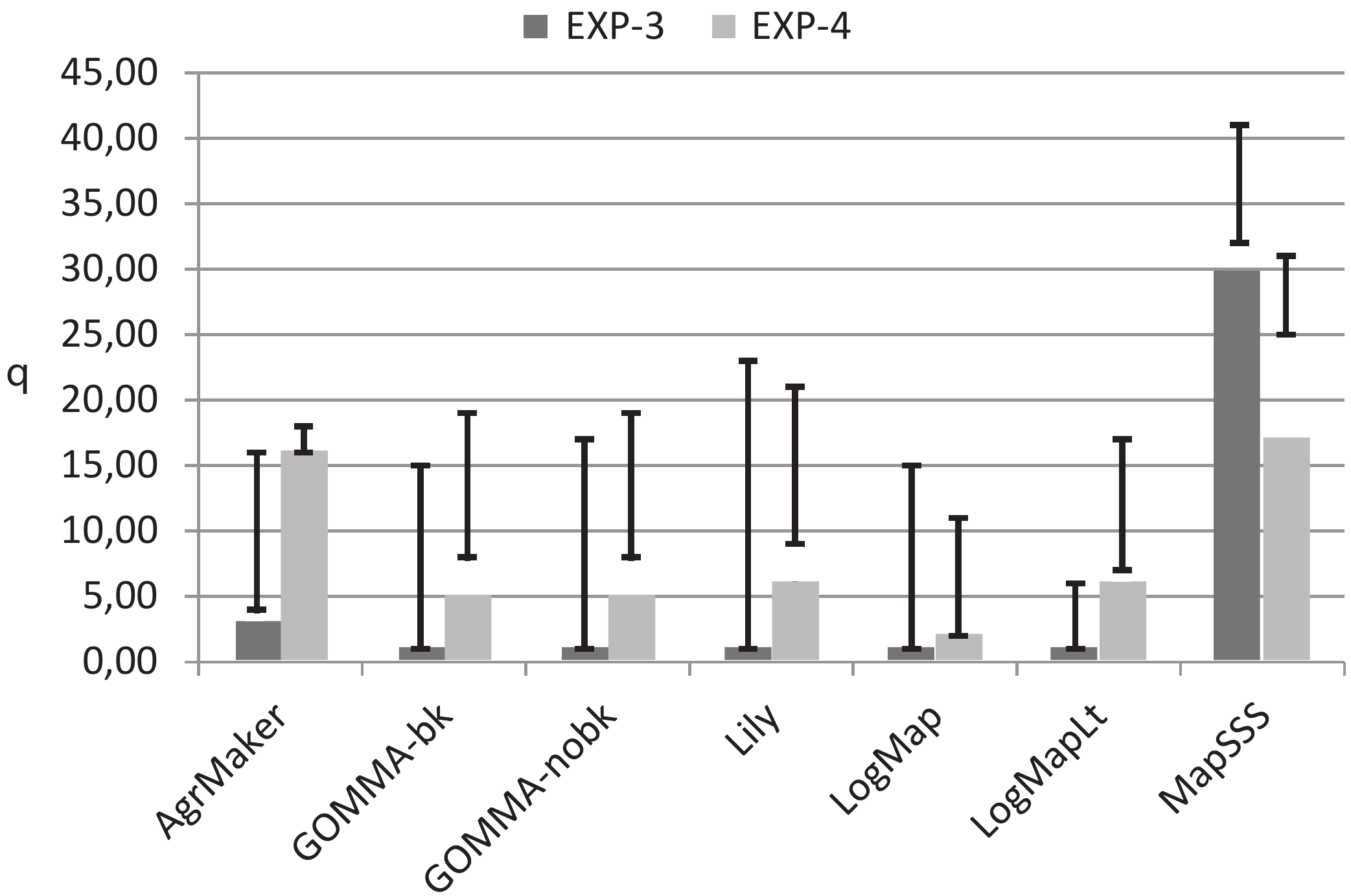}
    \label{fig:query_anatomy}
}
\vspace{-7pt}
\caption[]{The bars show the avg. number of queries ($q$) needed by RIO, grouped by matching tools. The distance from the bar to the lower (upper) end of the whisker indicates the avg. difference of RIO to the queries needed by the per-session better (worse) strategy of SPL and ENT, respectively.}
\label{fig:queries}
\end{figure}

Results of both experimental sessions, $\langle\text{EXP-1,EXP-2}\rangle$ and $\langle\text{EXP-3,EXP-4}\rangle$, are summarized in Figure~\ref{fig:query_conference} and Figure~\ref{fig:query_anatomy}, respectively. For the ontologies produced by each of the matching systems and for the different experimental scenarios, the figures show the (average) number of queries asked by RIO and the 
%distances to the average minimum and maximum number of queries needed by the per-session best and worst strategy, respectively. 
(average) differences to the number of queries needed by the per-session better and worse strategy of SPL and ENT, respectively. 
%Note that the minimum and maximum number of queries was calculated per debugging session.
%the best and worst strategy was variable over the sessions
The results illustrate clearly that the average performance achieved by RIO was always substantially closer to the better than to the worse strategy. 
In both EXP-1 and EXP-2, throughout 74\% of 27 debugging sessions, RIO worked as efficiently as the best strategy (Figure \ref{tab:best_strategies}).  
% which was 18 of 26 times the entropy method \textit{E}.
In more than 25\% of the cases in EXP-2, RIO even outperformed both other strategies; in these cases, RIO could save more than 20\% of user interaction on average compared to the best other strategy. In one scenario involving OWL-CTXmatch in EXP-1, it took ENT 31 and SPL 13 queries to finish, whereas RIO required only 6 queries, which amounts to an improvement of more than 80\% and 53\%, respectively. In $\langle\text{EXP-3,EXP-4}\rangle$, the savings achieved by RIO were even more substantial. RIO manifested superior behavior to both other strategies in 29\% and 71\% of cases, respectively. Not less remarkable, in 100\% of the tests in EXP-3 and EXP-4, RIO was at least as efficient as the best other strategy.
%
%This indicates the positive effect of the learning ability of RIO, which is able to adapt the strategy flexibly in order to operate more successfully.
%This is the positive effect of the learning ability of \textit{RIO}. 
Table~\ref{tab:time_query}, which provides the average number of queries per strategy, demonstrates that, overall, RIO is the best choice in all experiments. Consequently, RIO is suitable for both good meta information as in EXP-1 and EXP-3, where $\dt$ has high probability, and poor meta information as in EXP-2 and EXP-4, where $\dt$ is a-priori less likely. Additionally, Table~\ref{tab:time_query} illustrates the (average) overall debugging time assuming that queries are answered instantaneously and the reaction time, i.e. the average time between two successive queries. Also w.r.t. these aspects, RIO manifested good performance. Since the times consumed by either of the strategies in $\langle\text{EXP-1,EXP-2}\rangle$ are almost negligible, consider the more meaningful results obtained in $\langle\text{EXP-3,EXP-4}\rangle$. While the best reaction time in both experiments was achieved by SPL, we can clearly see that SPL was significantly inferior to both ENT and RIO concerning the user interaction required and the overall time. RIO revealed the best debugging time in EXP-4, and needed only $2.2\%$ more time than the best strategy (ENT) in EXP-3. However, if we assume the user being capable of reading and answering a query in, e.g., half a minute on average, which is already quite fast, then the overall time savings of RIO compared to ENT in EXP-3 would already account for $5\%$. Doing the same thought experiment for EXP-4, using RIO instead of ENT and SPL would save $25\%$ and $50\%$ of debugging time on average, respectively.
%In EXP-3 and EXP-4 there was a further matching system, namely AROMA, that we want to contemplate separately, since it constitutes an outlier in terms of user interaction involved and computation times. In EXP-3, RIO was the winner with an overall debugging time of 1179s compared to 1330s (ENT) and 1631s (SPL). The queries needed and reaction times were (36q,33s) for RIO, (32q,51s) for SPL and (47q,28s) for ENT. In EXP-4, ........................
All in all, the measured times confirm that RIO is well suited as an \emph{interactive} debugging approach.
%Additionally, we see that in the scenario where ontology axioms, which have low probability, are included in $\dt$ as well, $\dt$ has lower probability and is thus calculated later in the HS-Tree. As a consequence, it takes all the strategies longer to isolate $\dt$.........

For SPL and ENT strategies, the difference w.r.t. the number of queries per test run between the better and the worse strategy was absolutely significant, with a maximum of 2300\% in EXP-4 and averages of 190\% to 1145\% throughout all four experiments, measured on the basis of the better strategy (Figure~\ref{fig:best_diff}\subref{fig:diff_SPL_ENT}). 
Moreover, results show that the different quality of the prior fault probabilities in \{\text{EXP-1},\text{EXP-3}\} compared to \{\text{EXP-2},\text{EXP-4}\} clearly affected the performance of the ENT and SPL strategies (see first two rows in Figure~\ref{fig:best_diff}\subref{tab:best_strategies}). This perfectly motivates the application of RIO. 

%Furthermore, for each of the strategies ENT, SPL and RIO, we measured the average time needed for the entire debugging session, the average reaction time, i.e. time between two queries, and the average number of executed consistency checks (Table~\ref{tab:time_query}). Concerning computation time, ENT proved to be most efficient, followed by SPL and RIO. The reason why RIO took up more time is because of the incorporated search for the best alternative query and associated consistency checks. However, the average time RIO required from the confirmation of a query answer until the determination of a successive query, is about $700$ ms, which is still a good value.

%\begin{table}
%\centering
%\begin{tabular}{@{\extracolsep{0pt}} l|c|c|c|c|c }
%System  		& $|\mo_{i\Align j}|$ &	$|\Align_{ij}|$	& \#C/\#P 	& \#D/min/max 	& \#CS/min/max \\\hline
%COMA++ 			& 347									&		19.1					& 76 / 66		&	204 / 1.50 / 4.88		& 7 / 4.88 / 7.13 \\
%Falcon-AO 	& 312									&		21.3					& 76 / 56		&	71 / 1.33 / 3.33		& 5 / 5.00 / 6.67 \\
%HMatch 			& 377									&		24.1					& 78 / 77		&	13 / 1.33 / 3.67 		&	19 / 4.67 / 7.67 \\
%OWL-CTXmatch& 329									&		28.4					& 70 / 68		&	74 / 2.08 / 5.25		& 10 / 3.25 / 4.50 \\ \hline
%\end{tabular}
%%\vspace{-10pt}
%\caption{Average values of ontology metrics computed over a set of ontologies $\mo_{i\Align j}$ generated by a matching system. Expressivity of the ontologies: $\mathcal{SIN(D)}, \mathcal{SHIN(D)}, \mathcal{ALCIF(D)},
%\mathcal{ALCIN(D)}$}\label{tab:onto_metrics}
%%\vspace{-20pt}
%\end{table}

\vspace{-15pt}

\vspace{-10pt}
\begin{table}
\centering
\begin{tabular}{@{\extracolsep{0pt}} l|c|c|c||c|c|c||c|c|c||c|c|c}
		&  \multicolumn{3}{c||}{EXP-1}								  &  \multicolumn{3}{c||}{EXP-2}										&  \multicolumn{3}{c||}{EXP-3}					&  \multicolumn{3}{c}{EXP-4} \\
		 
		\hline
 		& debug					& react 			& $q$ 					& 	debug 				& react 			& $q$ 					& debug 				& react 				& $q$ 					&  debug				&  	react				& $q$	\\\hline
ENT &  1860					& 262 				& 3.67 					&  	1423					& 204 				& 5.26					& \textbf{60928}& 12367					& 5.86 					& 74463 				& 5629 					& 11.86	\\
SPL &  \textbf{1427}& \textbf{159}& 5.70 					&  	\textbf{1237}	& \textbf{148}& 5.44					& 104910				&	\textbf{4786}	& 19.43 				& 98647 				& \textbf{4781}	& 18.29	\\
RIO &  1592					& 286 				& \textbf{3.00}	&  	1749					&	245  				& \textbf{4.37}	& 62289					&	12825					& \textbf{5.43}	& \textbf{66895}& 8327 					& \textbf{8.14}	\\ \hline
\end{tabular}
\vspace{1pt}
\caption{Average time (ms) for the entire debugging session (debug), average time (ms) between two successive queries (react), and average number of queries ($q$) required by each strategy.} 
\vspace{-22pt}
\label{tab:time_query}
\end{table}

%\begin{table}
%\centering
%\begin{tabular}{@{\extracolsep{0pt}} l|c|c|c|c|c|c|c }
%System  	& $|\mo_{i\Align j}|$ & $|\Align_{ij}|$		& \#q       &  react 	&   cc    &    debug 	& \#D/min/max 		\\ \hline
%AgrMaker 	&		17818							& 1435						  &	  1				&		17761	&		1798	&			23092	&		12 / 4 / 29		\\ 
%Aroma   	&		17661							& 1278						  &	  12			&		20535	&		498 	&			265458&		48 / 9 / 14		\\ 
%GOMMA-bk 	&		17844							& 1461						  &	  1				&		11355	&		1309	&			13667	&		8 / 2 / 28		\\ 
%GOMMA-nobk&		17844							& 1461						  &	  1				&		11282	&		1309	&			13584	&		8 / 2 / 28		\\ 
%Lily		 	&		17751							& 1368						  &	  1				&		61651	&		3034	&			73049	&		10 / 2 / 55		\\ 
%LogMap	 	&		17772							& 1389						  &	  1				&		9482	&		1128	&			11445	&		4 / 1 / 27		\\ 
%LogMapLt 	&		17530							& 1147						  &	  1				&		10269	&		1235	&			12278	&		8 / 2 / 25		\\ 
%MapSSS  	&		17595							& 1212						  &	  3				&		18085	&		1004	&			62407	&		24 / 5 / 30		\\ \hline
%\end{tabular}
%%\vspace{-10pt}
%\caption{EXP-3 results: For explanations of column labels see Tables \ref{tab:onto_metrics} and \ref{tab:time_query}. }\label{tab:anatomy_results}
%%\vspace{-20pt}
%\end{table}